\documentclass{article}
\usepackage{ijcai24}
\usepackage{times}
\usepackage{soul}
\usepackage{url}
\usepackage[hidelinks]{hyperref}
\usepackage[utf8]{inputenc}
\usepackage[small]{caption}
\usepackage{graphicx}
\usepackage{amsmath}
\usepackage{amssymb}
\usepackage{amsthm}
\usepackage{booktabs}
\usepackage{algorithm}
\usepackage{algorithmic}
\usepackage{multirow}
\usepackage{subfigure}
\usepackage{csquotes}
\usepackage{xcolor}
\usepackage[switch]{lineno}

\newtheorem{example}{Example}

\title{How Can Large Language Models Understand Spatial-Temporal Data?}

\author{
    Lei Liu, Shuo Yu, Runze Wang, Zhenxun Ma, Yanming Shen \\
\affiliations
    Dalian University of Technology \\
\emails
    \{ll1719684334, shuo.yu, runze\_wang, orange, shen\}@mail.dlut.edu.cn
}

\begin{document}
\maketitle

\begin{abstract}
While Large Language Models (LLMs) dominate tasks like natural language processing and computer vision, harnessing their power for spatial-temporal forecasting remains challenging. The disparity between sequential text and complex spatial-temporal data hinders this application. To address this issue, this paper introduces STG-LLM, an innovative approach empowering LLMs for spatial-temporal forecasting. We tackle the data mismatch by proposing:
1) STG-Tokenizer: This spatial-temporal graph tokenizer transforms intricate graph data into concise tokens capturing both spatial and temporal relationships; 
2) STG-Adapter: This minimalistic adapter, consisting of linear encoding and decoding layers, bridges the gap between tokenized data and LLM comprehension. By fine-tuning only a small set of parameters, it can effectively grasp the semantics of tokens generated by STG-Tokenizer, while preserving the original natural language understanding capabilities of LLMs.
Extensive experiments on diverse spatial-temporal benchmark datasets show that STG-LLM successfully unlocks LLM potential for spatial-temporal forecasting. Remarkably, our approach achieves competitive performance on par with dedicated SOTA methods. 
% Code is available at https://anonymous.4open.science/r/09289449AB.

\end{abstract}

\section{Introduction}
\begin{figure*}[t]
    \centering
    \includegraphics[width=2.0\columnwidth]{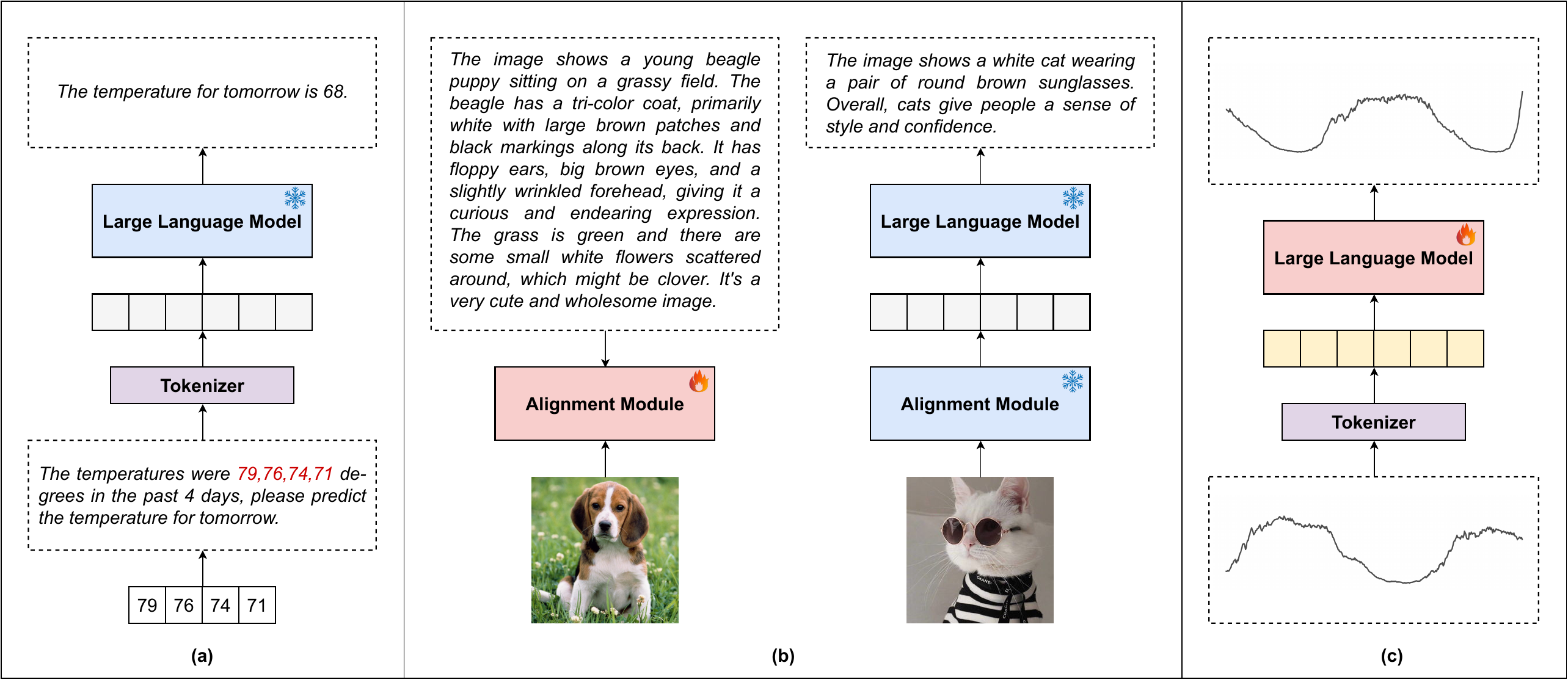}
    \caption{Possible solutions for applying LLMs to spatial-temporal forecasting, where \includegraphics[height=1em]{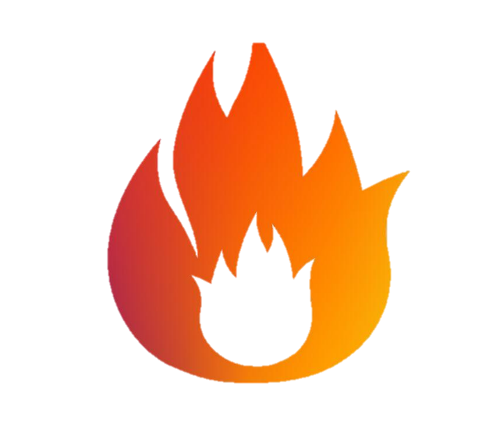} denotes trainable, and \includegraphics[height=1em]{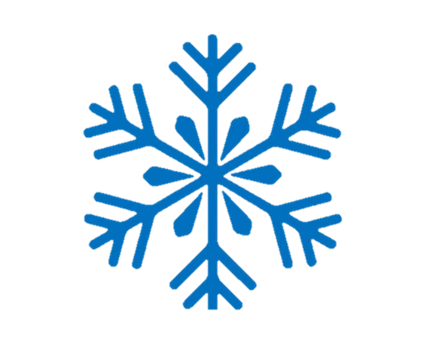} denotes frozen.}
    \label{alignment}
\end{figure*}

Spatial-temporal forecasting is a vital task in machine learning, with extensive applications ranging from traffic, weather, and to the spread of epidemics. Existing studies \cite{li2021spatial,lan2022dstagnn,rao2022fogs,jiang2023pdformer} have shown efficacy in capturing and modeling the intricate and dynamic spatial-temporal dependencies. However, their practical applications still suffer from issues like dedicated model design for a specific domain, data scarcity, and poor generalization, etc.

Large Language Models (LLMs) have demonstrated remarkable success in natural language processing \cite{radford2019language,touvron2023llama}, computer vision \cite{dai2305instructblip,berrios2023towards}, recommendation systems \cite{Zhang_2023,huang2023recommender}, and time series analysis \cite{zhou2023one,chang2023llm4ts,jin2023time,gruver2023large}, etc. If LLMs can be applied to spatial-temporal forecasting, we can unleash the remarkable reasoning capability of LLMs. First, LLMs are trained on massive data, giving them access to a vast amount of real-world knowledge and global patterns. This can be beneficial for understanding complex relationships in spatial-temporal forecasting and accounting for unforeseen events or contextual factors. Also, LLMs can be effective in situations with limited or sparse data, where traditional forecasting methods may struggle. Furthermore, LLMs can be adapted to a wide range of spatial-temporal forecasting tasks through prompting and fine-tuning. This allows for customization and tailoring the model to a specific domain.

However, LLMs are originally designed to process the sequential text data, instead of the spatial-temporal data. Therefore, {\it the key to achieving accurate spatial-temporal forecasting with LLMs lies in how to make LLMs comprehend spatial-temporal data.} 

One straightforward solution is to describe the spatial-temporal data using natural language \cite{xue2023promptcast,gruver2023large}, where the spatial-temporal data is converted to text, as shown in Figure \ref{alignment}(a). However, this approach needs a large amount of tokens to fully describe the spatial-temporal data. For example, to describe one-hour spatial-temporal traffic data in PEMS07 (883 nodes), at least 10596 tokens are required. To further describe tasks, network topology, and other information, more tokens will be needed. The large amount of tokens may exceed the context window of the existing LLMs, making it inapplicable. Additionally, existing LLMs do not have the ability to infer complex spatial-temporal dependencies based on natural language descriptions \cite{rimban2023challenges}.

Inspired by \cite{zhao2023gimlet,dai2305instructblip}, one can also utilize the paired data of the spatial-temporal data with the associated text. In this way, the spatial-temporal data can be mapped to text through the methods such as alignment modules, as shown in Figure \ref{alignment}(b). Although this approach is currently widely used in the field of computer vision, it is heavily dependent on the availability of large-scale high-quality text and non-text paired data. It is not suitable for spatial-temporal forecasting because spatial-temporal datasets are commonly characterized by small scales and lack of the associated text.

The practice of fine-tuning LLMs to align time series has gained popularity recently \cite{zhou2023one,chang2023llm4ts,jin2023time}. This approach entails transforming time series into a limited set of tokens that LLMs do not comprehend, and subsequently fine-tuning the LLMs to grasp the semantics of tokens, as shown in Figure \ref{alignment}(c). However, existing methods are mainly focused on addressing time series data, without the capability of accurately capturing spatial-temporal dependencies, which is crucial to spatial-temporal forecasting. Besides, the fine-tuning strategies used for time series are not applicable to spatial-temporal forecasting due to task discrepancy.

To address these challenges, in this paper, we propose STG-LLM, an LLM empowered spatial-temporal forecasting approach. With STG-LLM, we first design a spatial-temporal graph tokenizer (STG-Tokenizer), which treats each node in a spatial-temporal graph as a token. Each token will contain time series data for the corresponding node, as well as additional information, such as temporal semantics. In this way, spatial-temporal data can be transformed into a limited set of tokens, which can be naturally combined with prompts. Note that these tokens are not yet understood by the LLMs, therefore, we propose a simple and effective spatial-temporal graph adapter (STG-Adapter), which consists of a linear encoding layer and a linear decoding layer. The STG-Adapter equipped with an LLM can understand the semantics of tokens by fine-tuning a small amount of parameters, while preserving the LLMs' original natural language understanding capabilities. Benefiting from our proposed approach, LLMs can understand the spatial-temporal data and make accurate predictions. We summarize our main contributions as follows:

\begin{itemize}
    \item We employ LLMs for spatial-temporal forecasting, which can fully utilize the capabilities of LLMs, without the need for exquisite model designs as required by the traditional approaches. 
      
    \item We design a spatial-temporal graph tokenizer (STG-Tokenizer), which can convert spatial-temporal data into tokens that can later be understood by an LLM.
    
    \item We present a simple and effective spatial-temporal graph adapter (STG-Adapter), which can make an LLM understand the semantics of tokens by fine-tuning only a few parameters. 
    
    \item Experiments on spatial-temporal benchmark datasets demonstrate that STG-LLM can successfully make the spatial-temporal data understood by LLMs, and achieve performance comparable to the SOTA approaches.
\end{itemize}

\section{Related Work}
\subsection{Spatial-Temporal Forecasting}
Previous spatial-temporal forecasting studies involve statistical learning methods and conventional machine learning techniques such as VAR \cite{zivot2006vector} and SVR \cite{drucker1996support}. With the popularity of deep learning, \cite{li2021spatial,lan2022dstagnn,jiang2023pdformer} stack neural network modules to capture intricate spatial-temporal dependencies, which enables the models to comprehend and acquire spatial-temporal information at various hierarchical levels. Typically, these models consist of temporal dependency modules (e.g., CNN, RNN, and Transformer \cite{vaswani2017attention}) and spatial dependency modules (e.g., GCN, GAT \cite{velivckovic2017graph}, and Transformer). In addition, some methods \cite{li2021spatial,jiang2023pdformer} need to rely on rich prior knowledge and complex feature engineering to ensure performance. Nevertheless, these methods tend to excessively focus on improving complex spatial-temporal dependencies modeling capabilities, while still facing challenges in terms of domain adaptation, data scarcity resilience, and generalization.

\subsection{Large Language Models for Time Series}
With the rapid development of LLMs, they have been applied in the field of time series analysis. PromptCast \cite{xue2023promptcast} is among the first attempts to describe time series in text, and to utilize LLMs for reasoning. LLMTIME \cite{gruver2023large} encodes time series as a string of numerical digits, thereby transforming time series forecasting to next-token prediction. GPT4TS \cite{zhou2023one} utilizes a patching strategy to transform time series into tokens, followed by local fine-tuning to ensure that the semantics are comprehended by LLMs. Based on GPT4TS, LLM4TS \cite{chang2023llm4ts} utilizes a two-stage fine-tuning strategy to enhance the comprehension of token semantics in LLMs. TIME-LLM \cite{jin2023time} introduces a trainable reprogramming module instead of fine-tuning LLMs, which can better retain the original semantics of LLMs. However, existing methods are mainly focused on addressing time series data, without the capability of accurately capturing spatial-temporal dependencies, which is crucial to spatial-temporal forecasting. Besides, the fine-tuning strategies used for time series are not applicable to spatial-temporal forecasting due to task discrepancy.

\section{Problem Definition}
The spatial graph is represented as $\mathcal{G} = (\mathcal{V},\mathcal{E})$, where $\mathcal{V}$ is the set of nodes with $|\mathcal{V}|=N$, $\mathcal{E}$ is the set of edges with $|\mathcal{E}|=E$. Denote the observation of the graph at time step $t$ as $X_{t} = [x_{t}^{1}, x_{t}^{2}, ...,x_{t}^{N}] \in \mathbb{R}^{N\times F}$, where $x_{t}^{i} \in \mathbb{R}^{F}$ is the observation of node $i$ and $F$ is the number of features. In this way, we can represent previous $L$ time step observations as $X_{t-L+1:t}\in \mathbb{R}^{L\times N\times F}$, and the next $P$ time step observations as $X_{t+1:t+P}\in \mathbb{R}^{P\times N\times F}$. Given the spatial network $\mathcal{G}$, previous $L$ time step observations $X_{t-L+1:t}$, and the optional prompts $T_{p}$, the purpose of spatial-temporal forecasting with LLMs is to deduce the semantics and make next $P$ time step predictions $\hat{X}_{t+1:t+P}$ as close to the ground-truth $X_{t+1:t+P}$ as possible.

\begin{equation}\label{equation_1}
    [\mathcal{G},X_{t-L+1:t},T_{p}]\overset{LLM}{\rightarrow} X_{t+1:t+P}
\end{equation} 

\section{STG-LLM}
\begin{figure*}[t]
    \centering
    \includegraphics[width=2.0\columnwidth]{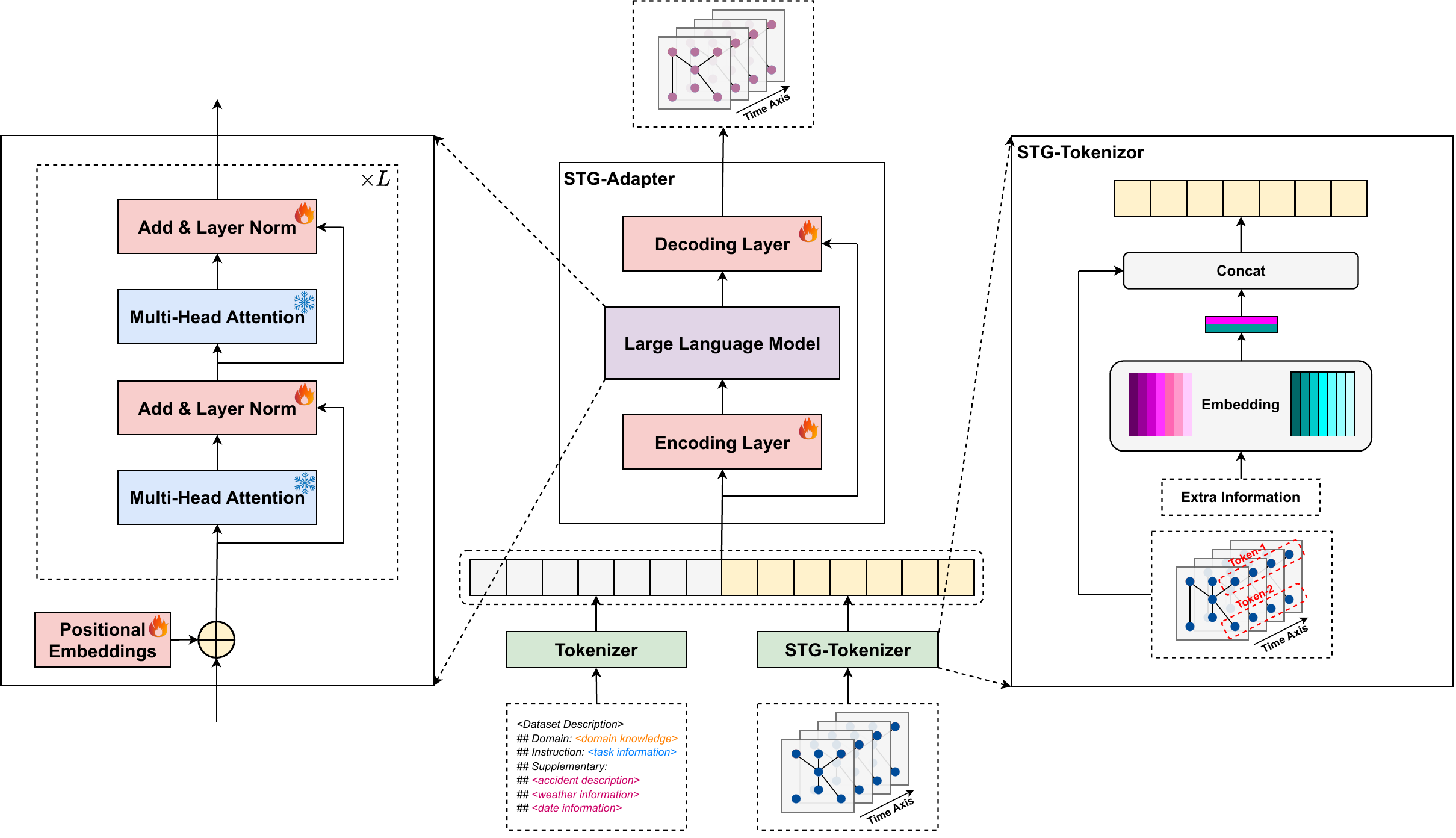}
    \caption{The overall framework of STG-LLM, where \includegraphics[height=1em]{figures/fire.png} denotes trainable, and \includegraphics[height=1em]{figures/snow.png} denotes frozen.}
	\label{stllm_framework}
\end{figure*}

In this section, we introduce our proposed STG-LLM. As shown in Figure \ref{stllm_framework}, STG-LLM mainly consists of a spatial-temporal graph tokenizer (STG-Tokenizer) and a spatial-temporal graph adapter (STG-Adapter). STG-Tokenizer is used to convert spatial-temporal data into concise tokens that can later be understood by an LLM. STG-Adapter equipped with an LLM can understand the semantics of tokens by fine-tuning a few parameters while preserving the LLMs's original natural language understanding capabilities. Next, we will introduce our proposed approach in details.

\subsection{Spatial-Temporal Graph Tokenizer}

Text-Tokenizer is a tool that processes sequential text into meaningful tokens \cite{radford2019language}, which helps LLMs understand text better, and improves the performance in various natural language processing tasks. Due to the significant disparity between spatial-temporal data and text data, the first step in using LLMs for spatial-temporal forecasting is to design a graph tokenizer that can convert spatial-temporal data into tokens. 

In order to achieve this goal, we design the spatial-temporal graph tokenizer (STG-Tokenizer). In simple terms, we treat each node as a token, so that we only need $N$ tokens to fully describe the spatial-temporal data, which reduces the number of tokens significantly. In this way, each token contains complete temporal semantic information, and the relationship between tokens can accurately express the spatial semantic information. Also, the spatial-temporal semantics in tokens can be easily understood by the LLMs. This is because that LLMs, e.g., GPT2, GPT3.5, and Llama, contain multi-head attention modules and feed forward modules, so that multi-head attention modules can effectively capture the spatial semantics between tokens and feed forward modules can effectively capture the temporal semantics within each token. Let $T_{g}=\{T_{g}^{1} , T_{g}^{2} ,...,T_{g}^{N} \} \in \mathbb{R}^{N\times L\times F} $ represent the tokens generated from the spatial-temporal data $X_{t-L+1:t}$, where $T_{g}^{i}\in  \mathbb{R}^{L\times F} $ denotes the token corresponding to node $i$.  

In order to make each token have richer semantics, STG-Tokenizer also involves an embedding module, which adds additional token-specific information to each token via embedding strategies. Take traffic flow forecasting as an example, we define time-of-the-day embeddings, TD-Embedding $B_{td}\in  \mathbb{R}^{K_{1}\times C_{1}}$ by dividing a day evenly into $K_{1}$ time steps. Similarly, we define day-of-the-week embeddings, DW-Embedding $B_{dw}\in  \mathbb{R}^{K_{2}\times C_{2}}$ by dividing a week equally into $K_{2}$ days. $C_{1}$ and $C_{2}$ denote the embedding dimension. We retrieve the TD-Embedding and the DW-Embedding respectively according to the last time step $t$ to obtain the time-of-the-day embeddings $E_{td}\in  \mathbb{R}^{N\times C_{1}}$ and day-of-the-week embeddings $E_{dw}\in  \mathbb{R}^{N\times C_{2}}$. Then, we can obtain the tokens $T_{e}=\{T_{e}^{1}, T_{e}^{2},...,T_{e}^{N}\} \in \mathbb{R}^{N\times (L\times F+C_{1}+C_{2})}$ with additional information, which is formulated as
\begin{equation}\label{equation_2}
    T_{e}^{i}= T_{g}^{i}\parallel E_{td}^{i} \parallel E_{dw}^{i},  
\end{equation}
where $\parallel$ means the concatenation operation.

\subsection{Spatial-Temporal Graph Adapter}
The LLMs, undergoing pretraining using extensive text corpora, are equipped with the capability of logical reasoning and a wealth of domain knowledge. These are helpful for spatial-temporal forecasting. However, the tokens generated by STG-Tokenizer are unfamiliar to LLMs. In order to make LLMs understand the tokens while retaining the LLMs's original natural language understanding capabilities, we propose the spatial-temporal graph adapter (STG-Adapter), which is a lightweight design, only consisting of a linear encoding layer and a linear decoding layer. 

First, the tokens generated by STG-Tokenizer will undergo a linear mapping process through an encoding layer, which is formulated as
\begin{equation}\label{equation_3}
    T_{q}=W_{1}T_{e}+b_{1}, 
\end{equation}
where $T_{q}\in R^{N\times D}$ denotes the tokens that have been processed by the encoding layer, $W_{1}\in \mathbb{R}^{(L\times F+C_{1}+C_{2})\times D} $, $b_{1} \in \mathbb{R}^{D}$, and $D$ denote a trainable weight matrix, bias, and the embedding dimensions of the LLMs, respectively.
The encoding layer serves two purposes: 1) It can keep the dimension of tokens consistent with the LLMs; and 2) It can capture the temporal semantics within the tokens to a certain extent.

For LLMs, it has been shown that well-crafted prompts can significantly improve their performance in various tasks. Our design can easily incorporate prompts, e.g., dataset information, weather information, traffic accidents, etc. Then, by putting together the tokens $T_{p}$ generated by Text-Tokenizer from prompts and the tokens $T_{q}$ generated by STG-Tokenzior from spatial-temporal data, the combined tokens $T$ to the LLM is  
\begin{equation}\label{equation_4}
    T = T_{p} \parallel T_{q},
\end{equation}
where $T\in R^{(M+N)\times D}$ denotes the combined tokens, $T_{p}\in R^{M\times D}$ denotes the tokens generated by Text-Tokenizer from prompts, and $M$ denotes the number of prompted tokens.

Then, we send the tokens $T$ to the LLM, which can be simply denoted as $H = LLM(T)$, where $H\in R^{(M+N)\times D}$ denotes the last hidden state of the LLM. An example of the combined tokens to the LLM is shown below, where the meanings of symbols such as \enquote{$\div$} and \enquote{$\triangleleft$} denote the tokens generated by STG-Tokenizer that are not yet understood by the LLM, and will be addressed by a lightweight fine-tuning.

\begin{example}\label{example1}
    Given the historical traffic values for 170 nodes from 10:55 to 11:50 on Monday. Your task is to predict the traffic values of next one hour. The historical traffic values of each node are as 
    $\eta$ $\div$ $\doteq$ $\updownarrow$ $\bullet$ $\longleftrightarrow$ $\triangleleft$ ... $\Omega$ $\circ$.
\end{example}

Moreover, we introduce a decoding layer to obtain predictions according to the spatial-temporal dependencies captured by the LLM. The decoding layer consists of two linear layers and a residual connection in order to guide the model to pay attention to the temporal semantics. The decoding layer is formulated as
\begin{equation}\label{equation_5}
    \hat{X}_{t+1:t+P} = W_{3}((W_{2}H+b_{2})+T_{e})+b_{3},
\end{equation}
where $\hat{X}_{t+1:t+P}\in \mathbb{R}^{N\times P}$ denotes the predictions of next $P$ time steps, $W_{2}\in \mathbb{R}^{D\times (L\times F+C_{1}+C_{2})}$ and $W_{3}\in \mathbb{R}^{(L\times F+C_{1}+C_{2})\times P}$ denote trainable weight matrices, $b_{2} \in \mathbb{R}^{L\times F+C_{1}+C_{2}}$ and $b_{3} \in \mathbb{R}^{P}$ denote trainable biases.

The logical reasoning and domain knowledge of LLMs lies in multi-head attention modules and feed forward modules \cite{zhou2023one}, and these modules have a large number of parameters. To preserve the LLMs' original natural language understanding capabilities and reduce the cost of fine-tuning, we choose to freeze them while fine-tuning. Besides, to enhance the ability of LLMs to capture spatial-temporal dependencies, we perform fine-tuning the positional embeddings and layer normalization layer, a widely adopted practice in various tasks \cite{houlsby2019parameter,lu2021pretrained,zhou2023one}.

Benefiting from our proposed STA-Tokenizer and STG-Adapter, LLMs can efficiently understand the spatial-temporal data. Also, note that with our design, one can choose any suitable LLM based on dataset, tasks, and resources.

\section{Experiments}
To comprehensively validate the effectiveness of STG-LLM, we
conduct extensive experiments and attempt to answer the following questions:

\begin{itemize}
    \item How does STG-LLM perform on various spatial-temporal forecasting tasks compared to existing approaches? (Overall Performance)
      
    \item Can STG-LLM generalize to various downstream datasets by using a small amount of data? (Few-shot Forecasting)
    
    \item Do prompts improve the performance of STG-LLM? (Prompt Effectiveness)
    
    \item What is the effect of each key component of STG-LLM? (Ablation Study)

    \item Can STG-LLM enjoy competitive performance by finetuning a small number of parameters and using the similar parameters count as compared to current approaches? (Parameter Efficacy)
\end{itemize}

\subsection{Experimental Setup}
\subsubsection{Datasets}
We verify the effectiveness and generalization of our proposed STG-LLM on six representative real-world spatial-temporal public datasets, including four traffic datasets (PEMS03, PEMS04, PEMS07, and PEMS08), an electricity dataset (Electricity), and a financial dataset (ExchangeRate). PEMS0X are collected from the Caltrans Performance Measurement System (PeMS) in every 30 seconds and aggregated into 5-minutes windows. They are widely used in previous spatial-temporal forecasting studies \cite{lan2022dstagnn,rao2022fogs,jiang2023pdformer}. Electricity records the electricity consumption in kWh every 1 hour from 2012 to 2014, which contains 321 clients. ExchangeRate collects the daily exchange rates of eight countries including Australia, British, Canada, Switzerland, China, Japan, New Zealand, and Singapore from 1990 to 2010. Electricity and ExchangeRate are widely used in previous multivariate time series forecasting studies \cite{shao2023exploring}. The statistics of the six datasets are summarized in Table \ref{dataset_detail}.

\begin{table}
    \tiny
    \centering
    \begin{tabular}{cccc}
        \toprule
        Datasets            & Sensors  & Time Steps  & Time Range           \\
        \midrule
        PEMS03              & 358      & 26,208       & 09/2018 - 11/2018    \\
        PEMS04              & 307      & 16,992       & 01/2018 - 02/2018    \\
        PEMS07              & 883      & 28,224       & 05/2017 - 08/2017    \\
        PEMS08              & 170      & 17,856       & 07/2016 - 08/2016    \\
        Electricity         & 321      & 26,304       & 07/2016 - 07/2019    \\
        ExchangeRate        & 8        & 7,588        & 01/1990 - 10/2010    \\
        \bottomrule
    \end{tabular}
    \caption{Dataset description and statistics.}
    \label{dataset_detail}
\end{table}

\subsubsection{Baselines}
We compare our proposed STG-LLM with the following baselines, categorized into three classes: 
1) The traditional spatial-temporal forecasting methods, including VAR \cite{zivot2006vector} and SVR \cite{drucker1996support};
2) The deep learning methods with predefined graph structure for spatial-temporal forecasting, including STGCN \cite{yu2017spatio}, DCRNN \cite{li2017diffusion}, STSGCN \cite{song2020spatial}, STGODE \cite{fang2021spatial}, STFGNN \cite{li2021spatial}, DSTAGNN \cite{lan2022dstagnn}, FOGS \cite{rao2022fogs}, and PDFormer \cite{jiang2023pdformer};
3)  The deep learning methods without predefined graph structure for spatial-temporal forecasting, including GraphWaveNet \cite{wu2019graph}, AGCRN \cite{bai2020adaptive}, and STNorm \cite{deng2021st}.

\subsubsection{Experiment Settings}
To make a fair comparison with the baselines, same as existing studies, we divide the training set, validation set, and test set in a 6:2:2 ratio, and use the previous 12 time step observations to make the next 12 time step predictions. 

Experiments are conducted under the environment with Ubuntu 20.04, 12th Gen Intel(R) Core(TM) i9-12900K CPU, and GeForce RTX 3090. We set a uniform set of hyper-parameters for all datasets. We use GPT2 as the LLM of STG-LLM. The layers of GPT2 we used are $3$. The dimensions of TD-Embedding and DW-Embedding are set to $64$. The training epoch is $200$ with a $50$ iterations patience early stop strategy on the validation dataset. The batch size is $64$. The optimizer is Adam. The learning rate is $1\times 10^{-3}$. The weight decay is $0.05$. The loss function is Huber Loss. 
% We use Mean Absolute Error (MAE), Root Mean Squared Error (RMSE), and Mean Absolute Percentage Error (MAPE) as the metrics in the experiments. 

\subsection{Experiment Results}
\subsubsection{Overall Performance}
\begin{table*}[tb]\centering
        \tiny
	\begin{tabular}{ccccccccccccccccc}
        \toprule
		\multirow{2}{*}{Method} &&                \multicolumn{3}{c}{\raisebox{0.5ex}{PEMS03}}             &&            \multicolumn{3}{c}{\raisebox{0.5ex}{PEMS04}}                &&                    \multicolumn{3}{c}{\raisebox{0.5ex}{PEMS07}}     &&                     \multicolumn{3}{c}{\raisebox{0.5ex}{PEMS08}}                    \\ \cline{3-5} \cline{7-9} \cline{11-13}  \cline{15-17}
			                    &	    &\raisebox{-0.5ex}{MAE}          &\raisebox{-0.5ex}{RMSE}          &\raisebox{-0.5ex}{MAPE(\%)}&             &\raisebox{-0.5ex}{MAE}          &\raisebox{-0.5ex}{RMSE}          &\raisebox{-0.5ex}{MAPE(\%)}&             &\raisebox{-0.5ex}{MAE}          &\raisebox{-0.5ex}{RMSE}          &\raisebox{-0.5ex}{MAPE(\%)}&             &\raisebox{-0.5ex}{MAE}          &\raisebox{-0.5ex}{RMSE}          &\raisebox{-0.5ex}{MAPE(\%)}\\ \midrule											
		        VAR             &       &23.65       &38.26         &24.51       &             &23.75            &36.66         &18.09       &             &101.20           &155.14          &39.69       &             &22.32            &33.83           &14.47        \\
                SVR             &       &21.97       &35.29         &21.51       &             &28.66            &44.59         &19.16       &             &32.97            &50.15           &15.43       &             &23.25            &36.15           &14.71        \\
                DCRNN           &       &18.18       &30.31         &18.91       &             &22.74            &36.58         &14.75       &             &23.63            &36.51           &12.28       &             &18.86            &28.18           &11.24        \\
                STGCN           &       &17.49       &30.12         &17.15       &             &21.76            &34.77         &13.85       &             &22.90            &35.44           &11.98       &             &17.84            &27.12           &11.21        \\
                GraphWaveNet    &       &19.12&32.77&18.89&             &24.89&39.66&17.29&             &26.39&41.50&11.97&             &14.56&25.65&12.15\\
                STSGCN          &       &17.48       &29.21         &16.78       &             &21.19            &33.65         &13.88       &             &24.26            &39.03           &10.20       &             &17.13            &26.79           &10.96        \\
                AGCRN           &       &15.98       &28.25         &15.23       &             &19.83            &32.26         &12.97       &             &22.37            &36.55           &9.12        &             &15.95            &25.22           &10.09        \\
                STFGNN          &       &16.77       &28.34         &16.30       &             &19.83            &31.87         &13.02       &             &22.07            &35.81           &9.21        &             &16.64            &26.21           &10.55        \\
                STGODE          &       &16.50       &27.84         &16.69       &             &20.85            &32.83         &13.78       &             &22.98            &36.19           &10.14\      &             &16.82            &26.24           &10.62        \\
                STNorm          &       &15.30&25.65&14.69&             &19.27&30.80&13.18&             &21.32&34.60&9.06&             &15.73&24.88&10.05\\ 
                DSTAGNN&       &15.57&27.21&\underline{14.68}&             &19.30&31.46&12.70&             &21.42&34.51&9.01&             &15.67&24.77&9.94\\
                FOGS&       &15.06&\textcolor{black}{\textbf{24.25}}&\textcolor{black}{\textbf{14.11}}&             &19.35&31.33&12.71&             &20.62&33.96&8.58&             &14.92&24.09&9.42\\
                PDFormer        &       &\textcolor{black}{\textbf{14.74}}&25.59&15.35&             &\underline{18.31}&\underline{29.97}&\textcolor{black}{\textbf{12.10}}&             &\underline{19.83}&\textcolor{black}{\textbf{32.87}}&\underline{8.53}&             &\textcolor{black}{\textbf{13.58}}&\underline{23.51}&\textcolor{black}{\textbf{9.05}}\\ \midrule
                STG-LLM           &       &\underline{14.97}&\underline{24.30}&16.00&             &\textcolor{black}{\textbf{18.14}}&\underline{30.11}&\underline{12.27}&             &\textcolor{black}{\textbf{19.82}}&\underline{33.06}&\textcolor{black}{\textbf{8.51}}&             &\underline{13.78}&\textcolor{black}{\textbf{23.15}}&\underline{9.10}\\ \bottomrule
    \end{tabular}
    \caption{Performance comparisons on traffic datasets, where the best results are in bold and the second best results are underlined.}
    \label{traffic_overall_performance}
\end{table*}

Table \ref{traffic_overall_performance} shows the prediction performance of our proposed STG-LLM and the baselines on traffic flow datasets. We can see that STG-LLM has surpassed most of the baselines in terms of performance. Only PDFormer's performance is close to that of our proposed STG-LLM. However, PDFormer stacks many neural network modules, such as semantic spatial attention, geographic spatial attention, temporal self-attention, and delay-aware feature transformation in order to improve performance. Besides, PDFormer also involves time-consuming feature engineering such as DTW matrix \cite{li2021spatial} and kShape clustering \cite{paparrizos2015k}. In contrast, STG-LLM neither uses complex model design nor involves complex feature engineering, but can still achieve similar performance to PDFormer. 

\begin{table}[t]\centering
        \tiny
	\begin{tabular}{cccccccc}
        \toprule
	\multirow{2}{*}{Method} & \multicolumn{3}{c}{\raisebox{0.5ex}{Electricity}}          &&              \multicolumn{3}{c}{\raisebox{0.5ex}{ExchangeRate}}            \\ \cline{2-4} \cline{6-8}
				           &\raisebox{-0.5ex}{MAE}          &\raisebox{-0.5ex}{RMSE}          &\raisebox{-0.5ex}{MAPE(\%)}&        &\raisebox{-0.5ex}{MAE}          &\raisebox{-0.5ex}{RMSE}          &\raisebox{-0.5ex}{MAPE(\%)}\\ \midrule
        GraphWaveNet          &233.72          &1877.55          &34.49        &        &0.0119      &0.0205       &1.42                \\
        AGCRN                 &\underline{224.75}          &\underline{1929.33}          &\underline{35.18}        &        &0.0150      &0.0247       &1.80                \\
        STNorm                &239.85          &2014.54          &33.65        &        &\underline{0.0118}      &\underline{0.0208}       &\underline{1.41}                \\ \midrule
        STG-LLM	              &\textcolor{black}{\textbf{214.11}}          &\textcolor{black}{\textbf{1822.55}}          &\textcolor{black}{\textbf{32.65}}        &        &\textcolor{black}{\textbf{0.0064}}      &\textcolor{black}{\textbf{0.0113}}       &\textcolor{black}{\textbf{0.85}}                \\ 
        \bottomrule
	\end{tabular}
	\caption{Performance comparisons on Electricity and ExchangeRate, where the best results are in bold and the second best results are underlined.}
	\label{other_overall_performance}
\end{table}

Table \ref{traffic_overall_performance} shows the prediction performance of our proposed STG-LLM and the baselines on Electricity and ExchangeRate. Note that for the Electricity and ExchangeRate datasets, there are no explicit graph structures. Among the spatial-temporal forecasting methods in baselines, only GraphWaveNet, AGCRN, and STNorm can be directly used for the spatial-temporal forecasting tasks without explicit spatial structure, and therefore, these three models are chosen as baselines. Table \ref{traffic_overall_performance} shows that our proposed STG-LLM is superior to GraphWaveNet, AGCRN, and STNorm in the spatial-temporal forecasting tasks without explicit spatial structure. 

According to Table \ref{traffic_overall_performance} and Table \ref{other_overall_performance}, we can make the following conclusions: 1) Different from existing methods, STG-LLM can achieve superior performance without prior knowledge (predefined graph structure, DTW matrix, Kshape clustering, etc.) and complex model design, which shows STG-LLM's strong ability to capture spatial-temporal dependencies; 2) STG-LLM has a strong generalization ability since it can be applied to the spatial-temporal forecasting tasks in various domains, such as traffic, electricity, and finance. Besides, users can choose the appropriate LLM based on the task, dataset, and resources.

\subsubsection{Few-shot Forecasting}
\begin{table}[t]
    \tiny
    \centering
    \begin{tabular}{ccccccccc}
        \toprule
        \multirow{2}{*}{Sample Size} & \multicolumn{3}{c}{\raisebox{0.5ex}{PEMS04$\to$PEMS08}} && \multicolumn{3}{c}{\raisebox{0.5ex}{PEMS08$\to$PEMS04}} \\
        \cline{2-4} \cline{6-8}
                               &\raisebox{-0.5ex}{MAE}          &\raisebox{-0.5ex}{RMSE}          &\raisebox{-0.5ex}{MAPE(\%)} &&\raisebox{-0.5ex}{MAE}          &\raisebox{-0.5ex}{RMSE}          &\raisebox{-0.5ex}{MAPE(\%)}\\
        \midrule
        0                      &44.78  &63.17  &31.75     &&37.87  &55.16  &29.91     \\
        20                     &19.59  &30.44  &14.14     &&24.33  &37.57  &18.98     \\
        50                     &18.62  &29.13  &13.13     &&22.97  &35.77  &17.12     \\
        100                    &17.86  &28.06  &12.37     &&21.96  &34.51  &15.97     \\
        200                    &16.53  &25.96  &11.25     &&21.06  &33.25  &14.96     \\
        500                    &15.39  &24.56  &10.28     &&19.68  &31.74  &13.41     \\
        1000                   &14.79  &24.05  &9.81      &&18.98  &30.92  &12.87     \\
        2000                   &14.32  &23.57  &9.48      &&18.57  &30.57  &12.87     \\
        \bottomrule
    \end{tabular}
    \caption{Performance of STG-LLM for few-shot forecasting on PEMS04 and PEMS08.\protect\footnotemark}
    \label{few_shot}
\end{table}

\footnotetext{For instance, the experiment involving PEMS04$\to$PEMS08 with a sample size of 100 indicates that we initially fine-tune STG-LLM on the full training data from PEMS04. Subsequently, we fine-tune STG-LLM again on PEMS08 using 100 samples.}

LLMs use large training datasets that cover a variety of topics, domains, and language styles, allowing them to learn language expression and logical reasoning. Therefore, LLMs can make use of context information to make up for the difficulties caused by data sparsity. STG-LLM, which utilizes LLMs, has acquired basic spatial-temporal dependent reasoning. This allows it to naturally solve other spatial-temporal forecasting tasks with limited data. 

To verify this, we conduct two sets of experiments. One is to learn the general spatial-temporal dependent reasoning ability by fine-tuning STG-LLM on the PEMS04, and then apply it to the PEMS08 with limited data. The other is to fine-tune STG-LLM on PEMS08, and then apply it to the PEMS04. Table \ref{few_shot} shows the performance of our proposed STG-LLM under the different sample sizes. We can see that STG-LLM performs poorly both on PEMS04 and PEMS08 without using sampled data (0\%). Different from tasks such as machine translation, sentiment analysis, and time series analysis, there is a large pattern difference between different spatial-temporal prediction datasets. For example, while fine-tuning STG-LLM on PEMS04 gives it general spatial-temporal dependent reasoning, it also adds PEMS04-specific spatial-temporal dependencies. In cases where the spatial-temporal dependencies of PEMS08 are different, there is a possibility of side effects occurring when using the STG-LLM containing PEMS04-specific spatial-temporal dependencies on PEMS08. The spatial-temporal dependence difference between PEMS04 and PEMS08 is analogous to the difference between machine translation and sentiment analysis tasks in natural language processing.

Fortunately, STG-LLM can familiarize the dataset-specific spatial-temporal dependencies with a small amount of data, and can derive good performance from its spatial-temporal dependent reasoning capabilities. According to Table \ref{traffic_overall_performance} and Table \ref{few_shot}, we can see the following results: 1) STG-LLM can achieve performance close to that of traditional deep learning methods for spatial-temporal forecasting with only 50 sample data on a new downstream dataset, which accounts for about 0.47\% and 0.49\% of the total data of PEMS08 and PEMS04 respectively; 2) By increasing the sampled data to 1000 (accounting for about 9.35\% and 9.82\% of the total data for PEMS08 and PEMS04, respectively), STG-LLM can outperform all of the baseline methods except PDFormer; 3) When the sampled data is 2000 (accounting for about 18.69\% and 19.64\% of the total data for PEMS08 and PEMS04, respectively), STG-LLM can achieve performance close to PDFormer and the STG-LLM using full training data of the new downstream dataset. These results illustrate that our proposed STG-LLM can generalize well to a new downstream dataset and is suitable for spatial-temporal forecasting tasks with data sparsity.

\subsubsection{Prompting Effectiveness}
Compared with the traditional deep learning methods, a prominent advantage of LLMs is their knowledge and reasoning ability acquired after massive pretraining. For example, given two prediction time periods from 14:00 to 15:00 on Sunday and 18:00 to 19:00 on Tuesday, LLMs are able to contain information that Tuesday is a working day, Sunday is a weekend, and 18:00 to 19:00 is the evening peak. Therefore, LLMs receive the prompt about the time period, enabling it to generate more accurate predictions. 

To verify the above hypothesis, we remove the temporal embedding module and tell the LLM the temporal information in prompts. The prompt is like {\it \enquote{Given the historical traffic values for 170 nodes from 10:55 to 11:50 on Monday. Your task is to predict the traffic values for the next one hour. The historical traffic values of each node are as follows.}}. As shown in Table \ref{performance_prompt}, STG-LLM can improve the prediction performance by adding the prompt with a time period. The experimental results directly show that the knowledge contained in LLMs can guide the spatial-temporal prediction results. Besides, it also proves that the tokens generated by STG-Tokenizer can be well combined with the prompts.

Also, prompt effectiveness is very practical in spatial-temporal forecasting. Taking traffic flow prediction as an example, if a traffic accident occurs on a certain road in the traffic network, we just need to add a prompt like {\it \enquote{The traffic accident occurred at node $i$ at time $j$, and the accident lasted for $t$ minutes.}} to LLMs, and then LLMs can guide predictions based on the knowledge like {\it \enquote{Traffic accidents may lead to less traffic flow in current node and an increase in traffic flow on other nodes.}} in LLMs.

\begin{table}[t]
\tiny
\centering
    \begin{tabular}{ccccc}\toprule
    Dataset                           & Method           & MAE          & RMSE           & MAPE(\%)          \\ \midrule
    \multirow{2}{*}{PEMS04}           & STG-LLM w/ prompt  &18.81         &30.62           & 12.73             \\ 
                                      & STG-LLM w/o prompt &19.26         &31.13           & 13.04             \\ 
    \multirow{2}{*}{PEMS08}           & STG-LLM w/prompt   &15.16         &24.13           & 9.67              \\ 
                                      & STG-LLM w/o prompt &15.71         &24.74           & 9.86              \\ \bottomrule
    \end{tabular}
    \caption{Prompt effectiveness, where STG-LLM w/ prompt (STG-LLM w/o prompt) denotes the STG-LLM with (without) temporal prompt, respectively.}
    \label{performance_prompt}
\end{table}

\subsubsection{Ablation Study}
To evaluate the effectiveness of each component in STG-LLM, we conduct ablation studies with 4 variants of STG-LLM as follows:
\begin{itemize}
    \item w/o LLM: it removes the LLM.
    \item w/o PE: it freezes the positional embeddings.
    \item w/o Tokenizer: it replaces the STG-Tokenizer with the tokenizer proposed in GPT4TS \cite{zhou2023one}.
    \item w/o Adapter: it removes the STG-Adapter, adopting padding to keep the input dimensions consistent with the LLM dimensions and a linear output layer to obtain the predictions.
\end{itemize}

\begin{table}[t]\centering
    \tiny
	\begin{tabular}{cccccccc}
		\toprule
		\multirow{2}{*}{Method} & \multicolumn{3}{c}{\raisebox{0.5ex}{PEMS04}}                  &&              \multicolumn{3}{c}{\raisebox{0.5ex}{PEMS08}}                   \\ \cline{2-4} \cline{6-8}
				           &\raisebox{-0.5ex}{MAE}          &\raisebox{-0.5ex}{RMSE}          &\raisebox{-0.5ex}{MAPE(\%)} &        &\raisebox{-0.5ex}{MAE}          &\raisebox{-0.5ex}{RMSE}          &\raisebox{-0.5ex}{MAPE(\%)} \\ \midrule		
	  w/o LLM	            &21.94           &35.12            &14.89           &        &16.46       &26.83        &11.07               \\
        w/o PE	              &20.51           &33.07            &14.13           &        &15.05       &24.42        &10.06               \\
        w/o Tokenizer         &26.91           &42.94            &18.04           &        &19.00       &31.83        &13.13               \\
        w/o Adapter	          &19.41           &31.50            &13.28           &        &15.95       &25.16        &11.13               \\
        STG-LLM	              &18.14           &30.11            &12.27           &        &13.78       &23.15        &9.10                \\
	\bottomrule
	\end{tabular}
    \caption{Ablation study on PEMS04 and PEMS08.}
    \label{ablation}
\end{table}

Table \ref{ablation} shows the comparative performance of these variants on PEMS04 and PEMS08. Based on the results, we can make the following conclusions: 1) Trainable positional embeddings play the role of location identification and determining the relationship between locations. It can assist the LLM in capturing spatial dependencies more efficiently; 2) The tokens generated by STG-Tokenizer are easier to be understood by LLMs because STG-Tokenizer makes each token contain complete semantic information and the relationship between tokens can accurately express the spatial-temporal dependencies; 3) The STG-Adapter equipped with an LLM can effectively understand temporal semantics within each token and the spatial semantics between tokens; 4) While the tokenizer proposed in GPT4TS is effective at making time series data understood by LLMs, it performs poorly in spatial-temporal forecasting tasks, especially in the tasks with strong spatial correlations, because the tokens generated by it cannot express the spatial dependencies in the spatial-temporal data.

\subsubsection{Parameter Efficacy}
To prove that STG-LLM only needs to fine-tune a few parameters to make LLMs understand the spatial-temporal data, Table \ref{parameter_efficay} shows the parameter usage of each module in STG-LLM. The results show that the trainable parameters of our proposed STG-Tokenizer and STG-Adapter are only 0.39\% (0.03\% and 0.36\% respectively) of the total parameters of STG-LLM. Most trainable parameters in the LLM are located in position embeddings. As shown in Table \ref{ablation}, STG-LLM can still achieve competitive performance without using position embeddings. In general, the trainable parameters in STG-LLM only account for 1.70\% of the total parameters, which is in the same order as traditional deep learning methods \cite{jiang2023pdformer}.

\begin{table}
    \tiny
    \centering
    \begin{tabular}{cccc}
        \toprule
        Method                & Trainable       & Ratio(\%)      \\
        \midrule
        STG-Tokenizer         &18,880            &0.03            \\
        STG-Adapter	          &217,640           &0.36            \\
        Position Embeddings	  &786,432           &1.29            \\
        Layer Norm	          &10,752            &0.02            \\ \midrule
        STG-LLM	              &1,033,704          &1.70            \\     
        \bottomrule
    \end{tabular}
    \caption{The number of parameters needs to be finetuned for each module in STG-LLM, where the total number of parameters is 60,885,480.}
    \label{parameter_efficay}
\end{table}

\section{Conclusion}
In this paper, we propose STG-LLM, an LLM empowered spatial-temporal forecasting approach. With STG-LLM, we first design a spatial-temporal graph tokenizer (STG-Tokenizer), which can convert the spatial-temporal data into a limited set of tokens. To make the tokens comprehended by LLMs, we then propose a simple and effective spatial-temporal graph adapter (STG-Adapter), which consists of a linear encoding layer and a linear decoding layer. The STG-Adapter equipped with an LLM can understand the semantics of tokens generated by STG-Tokenizer with fine-tuning a small amount of parameters, while preserving the LLMs' original natural language understanding capabilities. Experiments on spatial-temporal benchmark datasets demonstrate that LLMs can comprehend the spatial-temporal data and achieve state-of-art results. 

\bibliographystyle{named}
\bibliography{ijcai24}

\end{document}